\documentclass[sigconf,nonacm]{acmart}

\usepackage{graphicx}
\usepackage{subcaption}
\AtBeginDocument{%
  }

\setcopyright{acmlicensed}
\copyrightyear{2018}
\acmYear{2018}
\acmDOI{XXXXXXX.XXXXXXX}
\acmConference[Conference acronym 'XX]{Make sure to enter the correct
  conference title from your rights confirmation email}{June 03--05,
  2018}{Woodstock, NY}
\acmISBN{978-1-4503-XXXX-X/2018/06}

 \usepackage{multirow}
 \usepackage{enumitem} 
\acmSubmissionID{6155}
\newcommand{\locaname}{RF-CMG}
\begin{document}
\pagestyle{plain}
\title{Cross-Modal Generation: From Commodity WiFi to High-Fidelity mmWave and RFID Sensing}

\author{Zhixiong Yang}
\authornote{These authors contributed equally to this work.}
\affiliation{%
  \institution{Xinjiang University}
  \city{Urumqi}
  \country{China}
}

\author{Long Jing}
\authornotemark[1]
\affiliation{%
  \institution{Northwest University}
  \city{Xi'an}
  \country{China}
}

\author{Yao Li}
\affiliation{%
  \institution{Xinjiang University}
  \city{Urumqi}
  \country{China}
}

\author{Shuli Cheng}
\affiliation{%
  \institution{Xinjiang University}
  \city{Urumqi}
  \country{China}
}

\author{Guoxuan Chi}
\affiliation{%
  \institution{Tsinghua University}
  \city{Beijing}
  \country{China}
}

\author{Chenyu Wen}
\affiliation{%
  \institution{Northwest University}
  \city{Xi'an}
  \country{China}
}

\renewcommand{\shortauthors}{Yang and Jing, et al.}

\renewcommand{\shortauthors}{Trovato et al.}

\begin{abstract}
 AIGC has shown remarkable success in CV and NLP, and has recently demonstrated promising potential in the wireless domain. However, significant data imbalance exists across RF modalities, with abundant WiFi data but scarce mmWave and RFID data due to high acquisition cost. This makes it difficult to train high-quality generative models for these data-scarce modalities. In this work, we propose RF-CMG, a diffusion-based cross-modal generative method that leverages data-rich WiFi signals to synthesize high-fidelity RF data for scarce modalities including mmWave and RFID. The key insight of RF-CMG is to decouple cross-modal generation into high-frequency guidance and low-frequency constraint, which respectively learn high-frequency distribution from limited target modality data and preserve the underlying physical structure via low-frequency constraints during generation. On this basis, we introduce a Modality-Guided Embedding (MGE) module to steer the reverse diffusion trajectory toward the target high-frequency distribution, and a Low-Frequency Modality Consistency (LFMC) module to progressively enforce low-frequency constraints to suppress the accumulation of source-modality structural biases during inference, enabling high-quality target-modality generation. Performance comparison with several prevalent generative models demonstrates that RF-CMG achieves superior performance in synthesizing RFID and mmWave signals. We further showcase the effectiveness of the data generated by RF-CMG in gesture recognition tasks, and analyze the impact of the proportion of synthetic data on downstream performance.



\end{abstract}



\maketitle

\begin{figure}[htbp]
    \centering
     \includegraphics[width=\linewidth]{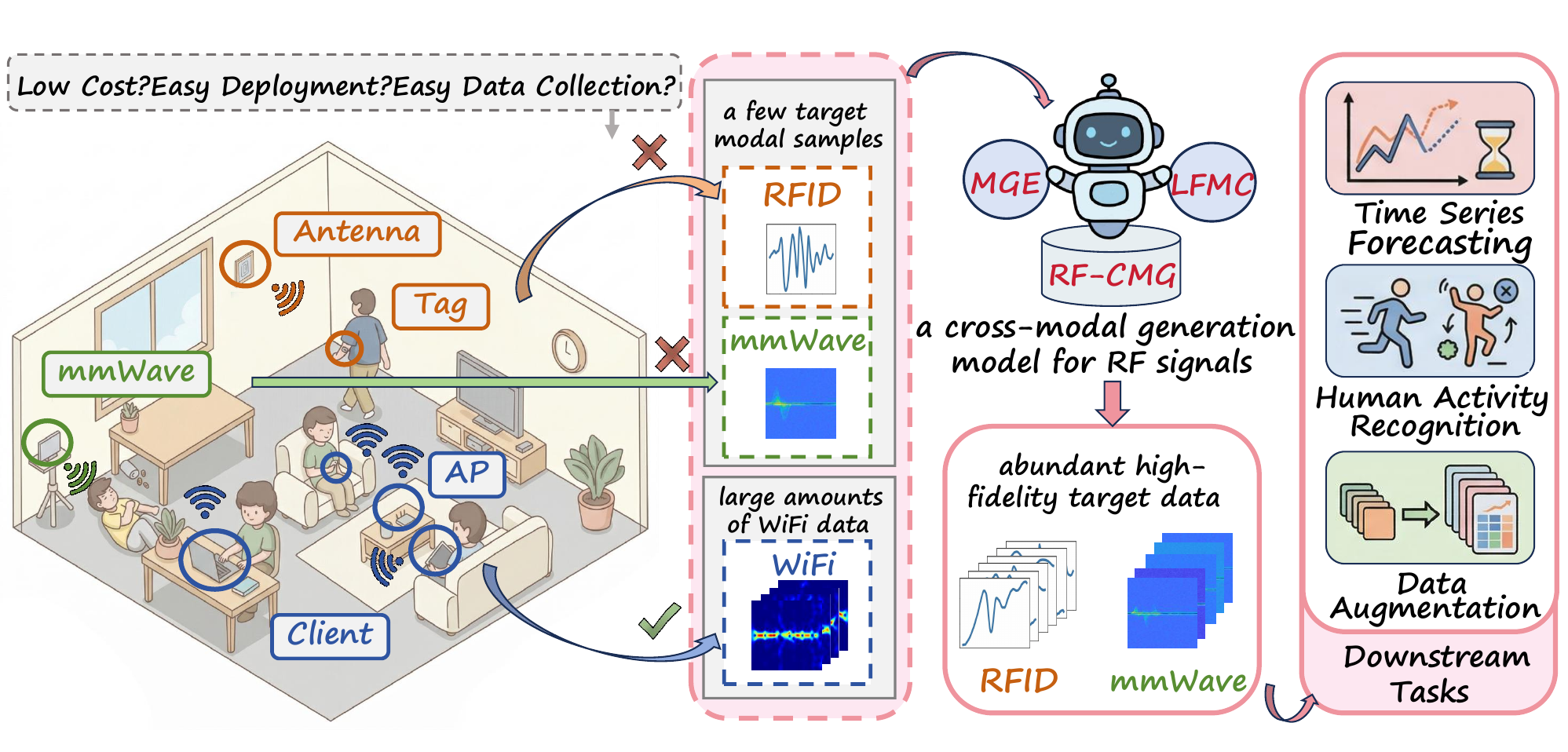}
    \caption{RF-CMG for cross-modal RF signal generation. By leveraging abundant low-cost WiFi data and a few target-modality samples, RF-CMG synthesizes high-fidelity mmWave and RFID signals for downstream wireless sensing tasks.}
    \label{fig:1}
\end{figure}
\section{Introduction}
Artificial Intelligence Generated Content (AIGC) has triggered a transformative impact in frontier fields such as computer vision (CV)~\cite{li2025back,chen2025nitrofusion,wang2025pixnerd} and natural language processing (NLP)~\cite{ustun2024aya,radford2018improving,takita2025systematic}, birthing a series of cutting-edge applications and products. Representative systems include Imagen~\cite{saharia2022photorealistic} and DALL·E~\cite{ramesh2022hierarchical} for image generation, as well as Gemini~\cite{team2023gemini} and ChatGPT~\cite{achiam2023gpt} for text generation, demonstrating the strong capability of AIGC in synthesizing high-fidelity data and improving generalization.

Nowadays, AIGC has demonstrated strong potential in wireless communication and sensing systems, enabling a variety of tasks such as data augmentation~\cite{chen2023rf}, channel estimation~\cite{zhou2025generative,chen2025generative}, and time-series prediction~\cite{hamdan2020variational,ren2025aigc}. While promising, training high-quality generative models for certain scarce modalities, such as millimeter-wave (mmWave) and Radio Frequency Identification (RFID), remains challenging due to the high deployment and data collection costs. In contrast, pioneering works in WiFi-based HAR have collected large-scale WiFi datasets covering diverse activities and made them publicly available~\cite{zhang2021widar3,zhang2018crosssense,zhai2021rise}, making WiFi an ideal source of training data. This naturally raises a compelling question: can low-cost source modalities be leveraged to synthesize high-cost target modalities through cross-modal generation? If feasible, such a paradigm could significantly reduce data acquisition costs in Fig.~\ref{fig:1} and unlock a scalable path toward wireless sensing systems.

Unfortunately, no system exists today that can realize such cross-modal generation. On one hand, generative models such as generative adversarial networks (GANs)~\cite{goodfellow2020generative}, variational autoencoders (VAEs)~\cite{kingma2013auto}, and diffusion models~\cite{ho2020denoising} provide strong capabilities in data synthesis~\cite{xu2025gansec,xiao2024diffusion}, domain adaptation~\cite{yan2025wi,wang2024xrf55}, and representation learning~\cite{yang2022rethinking}, yet they are largely limited to intra-modal generation. On the other hand, few-shot learning–based solutions~\cite{liu2024unifi,sheng2024metaformer} are used to transfer knowledge from a source modal to a target modal, but the significant modal gap caused by the different physical mechanisms of wireless signals makes it difficult for models to learn a complete cross-modal mapping. Most critically, when only a few target-modal samples are available, the model often captures only salient high-frequency patterns, such as specific multipath distributions or high-frequency noise, while overlooking the low-frequency structures that characterize the underlying physical space. This leads to an incorrect coupling between target-modal high-frequency features and source-modal low-frequency features, resulting in significant artifacts and distortions in the generated cross-modal data.

To address the aforementioned issues, we propose {\locaname}, the first \textbf{cross-modal generation} model for \textbf{RF} signals based on diffusion models. The innovation behind {\locaname} lies in its decoupling of the complex cross-modal generation process into two stages: \textit{High-Frequency Guidance} and \textit{Low-Frequency Constraints}. Such a decoupling is necessary under few-shot conditions because learning a complete cross-modal mapping is inherently difficult with limited target-modal samples.  Instead, the model is designed to capture the most discriminative fine-grained high-frequency patterns from scarce target data during training, and to further constrain the low-frequency components that characterize the underlying physical structure during generation. To translate our idea into practice, two key challenges must be addressed:


\begin{itemize}[leftmargin=*, itemsep=2pt, nosep]

\item \textbf{Model collapse occurs under scarce target-modality samples.} While target modality samples are scarce and fail to cover diverse distributions (e.g., variations in direction and posture), directly fine-tuning a pretrained source-modality diffusion model is prone to model collapse, thereby disrupting the generalizable priors learned in the source domain. In other words,  such collapse causes the model to lose its generative capability, fail to capture target-modality high-frequency characteristics, and generate invalid or uninformative outputs.

\item \textbf{Low-frequency structural bias accumulation.} During cross-modal generation, the generative prior of the source model continuously influences the state evolution at each reverse diffusion step, leading to the progressive accumulation of \textit{macroscopic structural biases} (e.g., spatial layout or global morphology) along the generation trajectory, thereby degrading the generation quality. Achieving consistent and stable low-frequency structural constraints throughout the generation trajectory without disrupting the pretrained generative prior remains challenging.

\end{itemize}

To address the first challenge, we incorporate an auxiliary MGE module to steer the generated outputs toward the target modality high-frequency distribution. Our intuition is that directly fine-tuning a pretrained source diffusion model  with scarce target modality samples can disrupt the integrity of the \textit{generation manifold}, thereby corrupting the pretrained generative prior. Therefore, rather than redefining the generative mapping, we freeze the source model to preserve the integrity of the generative manifold. We then introduce a Modality-Guided Embedding (MGE) to guide the reverse diffusion trajectory within the manifold toward the high-frequency distribution of the target modality, thereby generating its high-frequency features. For the second challenge, we formulate the problem as an optimization task. The objective is to impose low-frequency constraints on generated samples, maximizing the alignment of their macroscopic structures with a reference sample. We thus propose a Low-Frequency Modality Consistency (LFMC) module to progressively constrain the low-frequency components along the reverse diffusion trajectory, thereby suppressing the accumulation of source-modality structural biases and enabling high-quality target-modality generation.

Our key contributions are summarized as follows:
\begin{itemize}[leftmargin=*, itemsep=2pt, nosep]
\item We propose {\locaname}, the first cross-modal generative diffusion model tailored for RF signals. It enables the generation of scarce RF modalities, such as mmWave and RFID, with only a few target-modal samples by using large amounts of public WiFi data.

\item We introduce the MGE and LFMC modules, which decouple the complex cross-modal generation process into high-frequency guidance and low-frequency constraint, enabling controllable cross-modal generation. While our design and results are presented in the context of RF signals, the basic idea can be extended to other generative tasks.

\item We fully implement RF-CMG. Extensive evaluation results from case studies show RF-CMG’s efficacy.

\textbf{Community Contribution.} The code and pre-trained models of RF-CMG are publicly available. Our solution, in whole or in part, provides a practical toolkit for both academia and industry to advance AIGC in the RF domain. Moreover, by decoupling complex cross-modal generation into high-frequency guidance and low-frequency constraint, it enables fine-grained modeling of target modalities and offers a promising paradigm for stable generation in data-scarce scenarios across diverse modalities.
\end{itemize}

\section{Related Work}
We briefly review the related work in the following.

\textbf{Generative Wireless Sensing.} Generative Artificial Intelligence (GenAI) has emerged as a powerful paradigm in wireless sensing, enabling high-fidelity data synthesis and improved generalization. Existing RF data generation models can be broadly categorized into two types: environment modeling-based~\cite{wen2025wrf,lu2024newrf,zhang2026rf} and data-driven probabilistic generative models~\cite{chi2024rf,huang2023diffar,wang2024generative,mao2024wi,chen2022fidora}. The former construct 3D environment models from LiDAR or video and use ray tracing~\cite{mckown1991ray} to simulate RF signal propagation, enabling signal prediction at the receiver. However, they overlook the material and physical properties of targets that affect RF signal propagation. Recent work employs neural radiance fields for implicit modeling of RF-complex environments to estimate signal propagation~\cite{zhao2023nerf2,lu2024newrf,wen2025neural}, but requires a stationary receiver, hindering time-series data generation for wireless systems and human motion recognition. The latter employ GANs, VAEs, and diffusion models to learn and sample RF data distributions for dataset augmentation. Despite their strong generative capability, these approaches pose limitations due to training instability~\cite{liao2024tfsemantic}, high computational cost~\cite{burgess2018understanding}, and dependence on large-scale data~\cite{zhao2024airecg}. Most importantly, existing work is largely confined to intra-modal generation. Training high-quality generative models for certain modalities such as mmWave and RFID is difficult due to high deployment costs and large-scale data collection overhead. Motivated by this limitation, RF-CMG leverages low-cost WiFi as a source modality to learn a generative prior, enabling high-quality generation of scarce target modalities under limited target data constraints.

\begin{figure}[!t] 
  \centering
  \includegraphics[width=\linewidth]{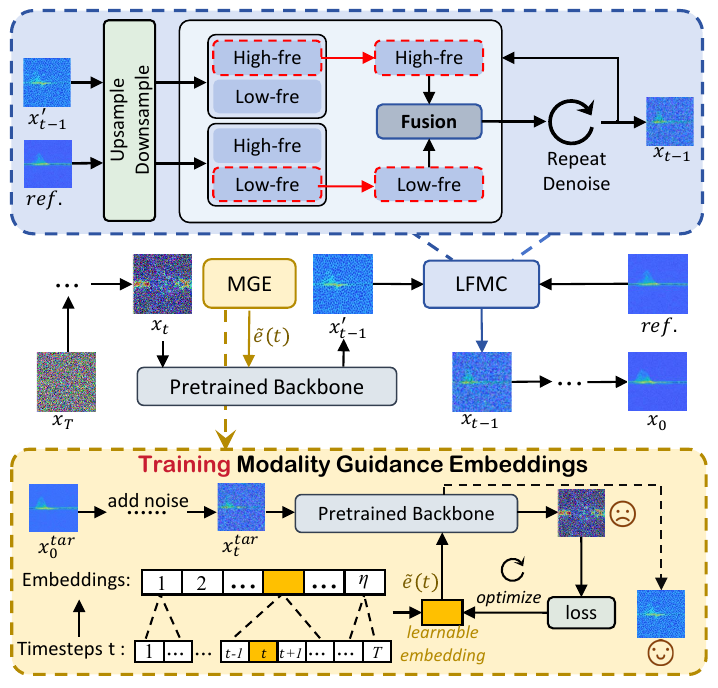}
  \caption{RF-CMG overview. A frozen WiFi-pretrained backbone is guided by MGE to learn target-modality high-frequency patterns, while LFMC imposes low-frequency structural constraints during the generation process.}
  \label{fig:teaser}
\end{figure}

\textbf{Few-shot Generation (FSG).} The objective of FSG is to adapt a pre-trained generative model to a target domain using limited samples, while preserving high-quality and diversity in the generated data~\cite{mondal2023few, wang2018transferring, zhao2022few}. Recent diffusion-based studies have shown that directly fine-tuning all model parameters under this setting easily causes overfitting and catastrophic forgetting, which has motivated a series of lightweight adaptation and guided sampling strategies~\cite{cao2024few}. Representative methods such as DDPM-PA~\cite{zhu2022few} and CRDI~\cite{cao2024few} improve few-shot adaptation by introducing sample-aware guidance and more stable inversion or reconstruction processes, thereby reducing the collapse commonly observed in naive fine-tuning. Meanwhile, guidance-based methods such as ILVR~\cite{choi2021ilvr} and Domain Guidance~\cite{zhong2025domain} steer the reverse diffusion trajectory with reference signals or domain-aware constraints, while more recent frameworks including the phasic content fusing diffusion model~\cite{hu2023phasic} and Uni-DAD~\cite{bahram2025uni} further emphasize preserving source-domain knowledge during adaptation to improve realism under extreme data scarcity. Despite this progress, these methods are still mainly developed for natural image generation\cite{lee2022progressive}, where the source and target domains remain visually aligned and share similar formation mechanisms. In cross-modal wireless sensing, however, the challenge is not only the lack of target samples, but also the fundamental mismatch in propagation physics, device characteristics, and low-frequency structural topology across modalities. Consequently, existing few-shot adaptation methods cannot directly resolve the severe source-target structural inconsistency in RF cross-modal transfer, which leaves this problem largely underexplored.
\section{Method}
\label{sec:method}
In this section, we offer a comprehensive introduction to each component of the RF-CMG framework, as illustrated in Fig.~\ref{fig:teaser}. The core idea of RF-CMG is to decouple the cross-modal generation process into two stages: high-frequency guidance and low-frequency constraint.

\subsection{Preliminaries}
\textbf{Denoising Diffusion Probabilistic Models(DDPMs)~\cite{ho2020denoising}.} DDPMs are a class of generative models that have shown remarkable performance in unconditional image generation tasks. It learns a data distribution by reversing a gradual Markovian noising process. The forward process adds Gaussian noise to the clean data $x_0$ over $t$ steps:
\begin{equation}
    q(x_t | x_{t-1}) = \mathcal{N}(x_t; \sqrt{1-\beta_t}x_{t-1}, \beta_t \mathbf{I}),
\end{equation}
where $\beta_t \in (0, 1)$ is a fixed variance schedule. Using the reparameterization trick, any intermediate state $x_t$ can be sampled directly from $x_0$:
\begin{equation}
    q(x_t | x_0) = \mathcal{N}(x_t; \sqrt{\bar{\alpha}_t}x_0, (1-\bar{\alpha}_t)\mathbf{I}),
\end{equation}
where $\alpha_t = 1 - \beta_t$ and $\bar{\alpha}_t = \prod_{s=1}^t \alpha_s$. The reverse process models the intractable conditional distribution $q(x_{t-1} | x_t)$ via a parameterized neural network $\boldsymbol{\epsilon}_\theta(x_t, t)$, which predicts the noise added to $x_t$. The network is trained using the simplified objective:
\begin{equation}
    \mathcal{L}_{DDPM} = \mathbb{E}_{t, x_0, \boldsymbol{\epsilon}} \left[ \| \boldsymbol{\epsilon} - \boldsymbol{\epsilon}_\theta(\sqrt{\bar{\alpha}_t}x_0 + \sqrt{1-\bar{\alpha}_t}\boldsymbol{\epsilon}, t) \|^2 \right].
    \label{eq:xt-1_pred}
\end{equation}
During inference, from the noise prediction formulation, 
the clean signal can be estimated as:
\begin{equation}
    \hat{x}_0(x_t, t) = \frac{x_t - \sqrt{1-\bar{\alpha}_t}\boldsymbol{\epsilon}_\theta(x_t, t)}{\sqrt{\bar{\alpha}_t}}.
    \label{eq:x0_pred}
\end{equation}

\begin{figure}[!t]
    \centering
    \includegraphics[width=\linewidth]{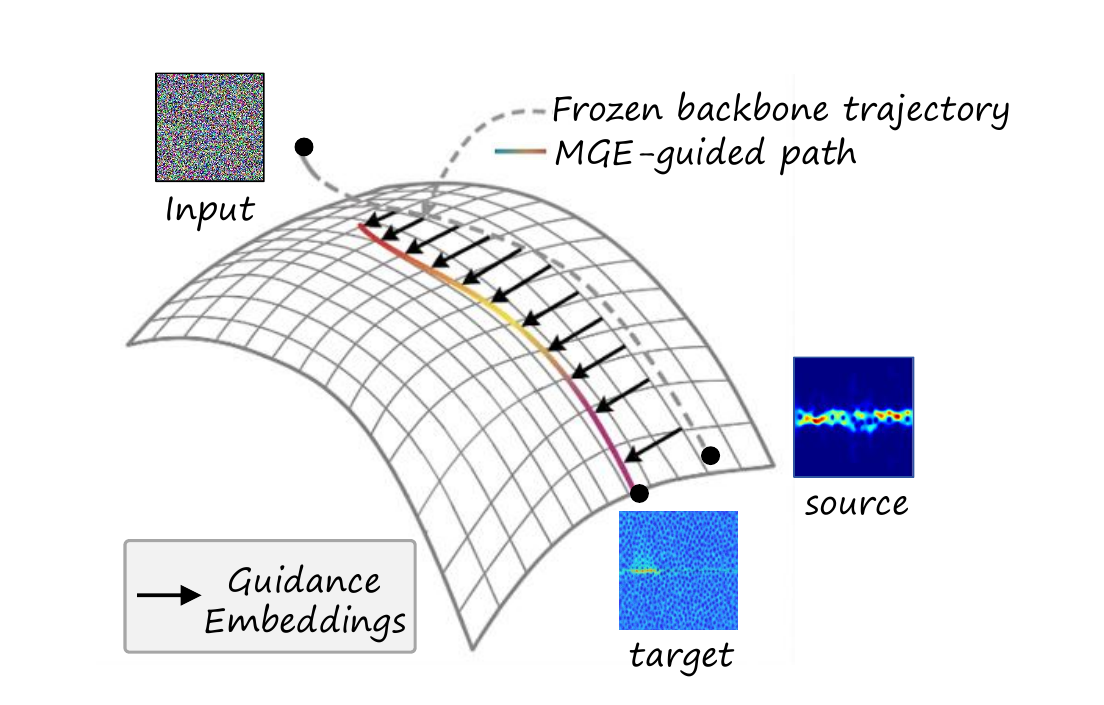}
    \caption{Manifold-space illustration of modality-guided embedding.}
    \label{fig:mge_manifold}
\end{figure}

\subsection{Modality Guided Embedding}
\label{subsec:guidance_embedding}
\textbf{Problem Formulation.} To be concrete, let $\mathcal{D}_S$ denote the source domain with abundant samples and $\mathcal{D}_T$ denote the target domain with only few samples. Assume a DDPM parameterized by $\theta$ is pre-trained on $\mathcal{D}_S$ to capture the spatial priors of RF sensing signals. Our aim is to use $\mathcal{D}_S$  to build a cross-modal generative model that synthesizes data in the scarce target modality. However, directly fine-tuning $\theta$ on $\mathcal{D}_T$ is prone to model collapse, which can disrupt the generalized prior learned from the source domain $\mathcal{D}_S$ and compromise the integrity of the generation manifold.

To this end, we introduce an auxiliary MGE module, denoted by $\mathbf{e}$, to steer the generation trajectory of the source modality toward the high-frequency distribution of the target modality. Specifically, we freeze the backbone of the source model to preserve the learned generative prior, which defines the underlying generation manifold as illustrated in Fig.~\ref{fig:mge_manifold}. On this basis, MGE applies timesteps-dependent corrections during reverse diffusion, which remain constrained within the fixed manifold of the source model. These corrections primarily affect the fine-grained details of the generated sample rather than the global structure. Consequently, MGE steers generation toward the high-frequency characteristics of the target modality without modifying the backbone parameters.



We divide the learning of MGE into two components: relaxed optimization during training and diversity-enhanced scheduling during inference.

\textbf{Relaxed Optimization During Training.}\label{3.2.1}
We adopt a \textit{conditional relaxed reconstruction strategy} that guides predicted states rather than strictly aligning noise. Specifically, we parameterize MGE as a piecewise time-dependent embedding $\mathbf{e}(t)$, where $t \in [1, T]$ denotes the diffusion timestep, to guide the multi-step reverse diffusion process. The timestep domain $[1, T]$ is partitioned into $\eta$ intervals, each assigned an independent learnable embedding, where a larger $\eta$ enables finer-grained modulation across diffusion steps. Within each interval, the model predicts both $\hat{x}_{t-1}$ and $\hat{x}_0$, where the former is optimized to align with the noised target state to guide local transitions, and the latter ensures global consistency. This design avoids reconstructing the entire generation mapping from scratch, and instead progressively corrects residual errors along a plausible reverse trajectory.

At each timestep $t$, the embedding is optimized via gradient descent according to the following objectives:
\begin{equation}
    \mathcal{L}_{MGE} =
    \| x^{tar}_{0} - \hat{x}_0 \|^2
    +
    \| x^{tar}_{t-1} - \hat{x}_{t-1} \|^2,
\end{equation}
where $x^{tar}_{0}$ is the ground-truth clean target sample, and $x^{tar}_{t-1}$ is the exact forward-diffused state of $x^{tar}_{0}$ at timestep $t-1$. This relaxed objective enables the learning of high-frequency characteristics of the target modality while preserving a valid diffusion trajectory. The prediction can be expressed as:
\begin{equation}
\hat{x}_0
=
F_\theta(x_t, t)
+
\Delta_\theta(x_t, t; \mathbf{e}(t)),
\end{equation}
where $x_t$ is the noisy input at timestep $t$, $F_\theta(x_t, t)$ denotes the source-domain prior prediction without modality guidance, and $\Delta_\theta(x_t, t; \mathbf{e}(t))$ is the embedding-induced corrective term parameterized by $\mathbf{e}(t)$.

Under few-shot supervision, learning a residual correction is substantially easier than relearning the entire generative mapping. Moreover, since this correction is constrained by the frozen source backbone, it primarily captures fine-grained high-frequency components of the target modality. Therefore, MGE introduces only a bounded adjustment to the source-domain generation trajectory, while the pre-trained diffusion model continues to determine the admissible generation dynamics and preserve the underlying structural prior.

\textbf{Diversity-Enhanced Inference.}
As mentioned in Section~\ref{3.2.1}, increasing $\eta$ improves reconstruction fidelity but reduces generation diversity in the target modality. To address this issue, we introduce a diversity-enhanced inference strategy based on an annealing schedule function, which perturbs the conditional embeddings with controlled noise during inference. The function perturbs it conditioned on the given input $\mathbf{e}(t)$:


\begin{equation}
\tilde{\mathbf{e}}(t) =
\sqrt{\gamma(t)}\mathbf{e}(t)
+
\nu \sqrt{1 - \gamma(t)} \mathbf{z},
\end{equation}
where $\mathbf{z} \sim \mathcal{N}(\mathbf{0}, \mathbf{I})$ is standard Gaussian noise, $\nu$ is the initial noise scale, and $\gamma(t)$ is defined as:
\begin{equation}
    \gamma(t) = \begin{cases}
    1 & t \le \beta \\
    \frac{\beta - t}{\beta - \alpha} & \beta < t < \alpha \\
    0 & t \ge \alpha
    \end{cases}
\end{equation}

Here, $\alpha$ and $\beta$ define the bounds of the noise scaling interval. As reverse diffusion proceeds from $t = T$ to $t = 0$, this schedule modulates the perturbation along the generation trajectory. At early stages ($t \ge \alpha$), $\gamma(t)=0$, which maximizes perturbation and encourages broader exploration in the embedding residual space. As the process enters $\beta < t < \alpha$, the perturbation is gradually reduced. At late stages ($t \le \beta$), $\gamma(t)=1$, which removes the perturbation. This schedule preserves trajectory stability while allowing controlled variation in fine-grained modality-dependent patterns, ultimately recovering the high-frequency fidelity governed by $\mathbf{e}(t)$.

\subsection{Low-Frequency Modality Consistency}
Although the MGE module enables the model to learn the high-frequency distribution of the target modality, its low-frequency components remain dominated by the source-domain $\mathcal{D}_S$ prior encoded in the frozen diffusion backbone $\boldsymbol{\epsilon}_\theta$. This source-driven macro-structural bias progressively accumulates along the reverse diffusion trajectory, ultimately leading to cross-modal structural mismatch rather than simple generation failure. 

To suppress such bias without altering the learned reverse diffusion process, we introduce a LFMC module that progressively refines the diffusion trajectory using a downsampled reference image. Let $\phi_N(\cdot)$ denote a low-pass filter implemented by downsampling and upsampling with scale factor $N$, which extracts global structure while maintaining dimensionality. During inference, given a reference sample $r \in \mathcal{D}_T$, we ensure that the low-frequency component of the generated sample $x_0$ matches that of the downsampled $\phi_N(r)$:
\begin{equation}
\phi_N(x_0) = \phi_N(r),
\end{equation}
Rather than modeling this constrained distribution explicitly, we enforce the consistency at each reverse diffusion step. At timestep $t$, we first use DDPM to compute the proposal distribution of $x'_{t-1}$:
\begin{equation}
    x'_{t-1} 
    \sim 
    p_\theta(x'_{t-1} \mid x_t, \tilde{\mathbf{e}}(t)),
\end{equation}
where $\tilde{\mathbf{e}}(t)$ denotes the perturbed modality embedding, while the backbone remains frozen. Let $r_{t-1}$ denote the reference sample corrupted to timestep $t-1$ by the forward diffusion process. We then perform a low-frequency projection, where $\mathbf{I}$ denotes the identity operator, and $(\mathbf{I}-\phi_N)$ corresponds to the complementary high-frequency residual:
\begin{equation}
    x_{t-1}
    =
    \phi_N(r_{t-1})
    +
    (\mathbf{I} - \phi_N)(x'_{t-1}),
    \label{eq:spectral_fusion}
\end{equation}
Eq.~\ref{eq:spectral_fusion} can be viewed as a frequency decomposition. The term $\phi_N(r_{t-1})$ anchors the macroscopic structure to the target modality, while $(\mathbf{I}-\phi_N)(x'_{t-1})$ maintains high-frequency details of the target modality synthesized under MGE guidance.

Importantly, LFMC does not modify the frozen backbone's reverse dynamics; it only constrains each transition in low-frequency space. As a result, generation is frequency-decoupled: the backbone defines admissible trajectories, MGE guides high-frequency adaptation, and LFMC progressively refines source-modality structural bias accumulation by enforcing cross-modal low-frequency consistency.



\subsection{Case Study}
\label{subsec:case_study}
In this section, two case studies validate the effectiveness of frequency decoupling theory and highlight the necessity of leveraging source-domain priors for few-shot cross-modal generation.


\begin{figure}[!t]
    \centering
    \includegraphics[width=\linewidth]{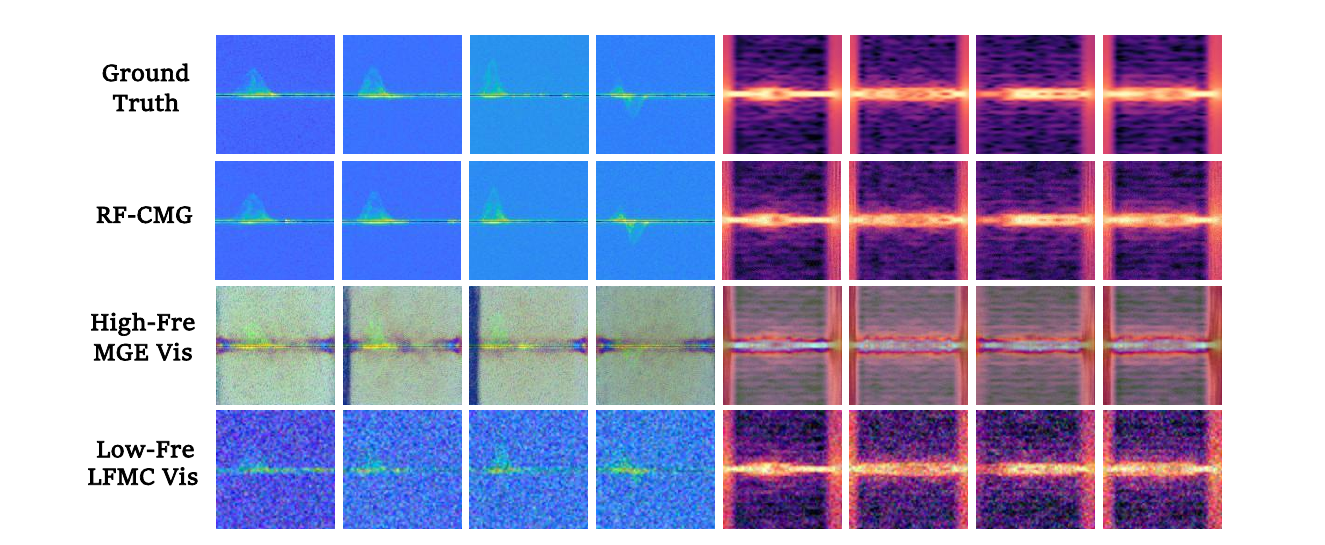}
    \caption{Visualization of the frequency decoupling mechanism.}
    \label{fig:mechanism_vis}
\end{figure}

\textbf{Mechanism Visualization: High-Low Frequency Decoupling.}
\label{subsubsec:mechanism_vis}
To validate the effectiveness of the MGE and LFMC modules in frequency decoupling, we present visualizations of their intermediate results during the generation process in Fig.~\ref{fig:mechanism_vis}. Specifically, the high-frequency saliency guided by the MGE (third row) strictly concentrates on the unique physical characteristics of the target modalities. For mmWave, this intuitively manifests as the granular micro-Doppler fringes and sparse multipath speckles typical of its spectrograms, whereas for RFID, it captures the dense phase fluctuations and environmental backscatter noise inherent to the hardware mechanism. Conversely, the macroscopic structure preserved by the LFMC (fourth row) accurately isolates the underlying low-frequency information of the samples, ensuring that the global spatial layout is effectively retained. This clear visual separation perfectly aligns with our theoretical design, proving that the proposed modules successfully achieve High-Low Frequency decoupling.


\begin{figure}[!t]
    \centering
     \includegraphics[width=\linewidth]{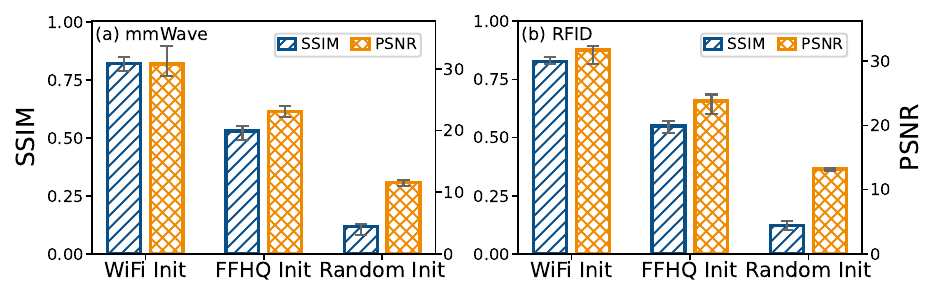}
    \caption{Comparison of cross-modal generation results using different pre-trained weights.}
    \label{fig:prior_necessity}
\end{figure}

\textbf{Necessity of Source Modality Prior.}
\label{subsubsec:source_prior}
When target-modality data are scarce, we argue that a source model pretrained on WiFi data provides a critical shared generative manifold for the target RF modality. To validate the necessity of this prior in preventing model collapse under few-shot conditions and ensuring generation fidelity, we evaluate our framework against random initialization and irrelevant prior initialization using the FFHQ image dataset across different target modalities. The results in Fig.~\ref{fig:prior_necessity} demonstrate that only the aligned WiFi prior can consistently support stable and meaningful generation across target modalities. Without such a prior, the model either collapses into noisy and unstructured outputs or produces visually plausible yet physically inconsistent patterns due to the mismatch between the learned source manifold and the RF sensing topology.



\begin{table*}[!t]
  \centering
  \caption{Quantitative comparison of generation quality across intra-modal (WiFi to WiFi) and cross-modal (WiFi to mmWave and RFID) scenarios.}
  \label{tab:main_results}
  \setlength{\tabcolsep}{2.5pt}
  \begin{tabular*}{\textwidth}
{c c c c c c c c c c c c c}
    \toprule
    \multirow{2}{*}{Method} & \multicolumn{4}{c}{WiFi $\rightarrow$ WiFi} & \multicolumn{4}{c}{WiFi $\rightarrow$ mmWave} & \multicolumn{4}{c}{WiFi $\rightarrow$ RFID}\\
    \cmidrule(lr){2-5} \cmidrule(lr){6-9} \cmidrule(lr){10-13}
    & FID $\downarrow$ & SSIM $\uparrow$ & PSNR $\uparrow$ & Intra-LPIPS $\uparrow$ 
    & FID $\downarrow$ & SSIM $\uparrow$ & PSNR $\uparrow$ & r-LPIPS $\rightarrow 1$ 
    & FID $\downarrow$ & SSIM $\uparrow$ & PSNR $\uparrow$ & r-LPIPS $\rightarrow 1$\\
    \midrule
    DDPM~\cite{ho2020denoising} & 27.08 & 0.83 & 27.15 & 0.42 & - & -& - & -& -& - & - & - \\
    Flow Matching~\cite{lipman2022flow} & 25.88 & 0.83 & 28.21 & 0.37 & - & -& -& -& - & - & - & - \\
    RF-Diffusion~\cite{chi2024rf} & \textbf{7.83} & 0.81 & 24.36 & 0.29 & - & -& - & -& -& - & - & - \\
    RICK~\cite{zhao2023exploring} & 117.51 & 0.55 & 17.24 & 0.26 & 275.34 & \textbf{0.83} & 27.31 & 1.80 & 271.22 & 0.56 & 21.69 & 1.67 \\
    CRDI~\cite{cao2024few} & 27.08 & 0.83 & 27.15 & 0.42 & 254.85 & 0.31 & 20.02 & 2.31 & 240.03 & 0.47 & 21.77 & 1.93 \\
    DoGFit~\cite{bahram2026dogfit} & 27.08 & 0.83 & 27.15 & 0.42 & 296.29 & 0.34 & 22.07 & 2.28 & 253.93 & 0.35 & 19.34 & 1.87 \\
    Uni-DAD~\cite{bahram2025uni} & 27.08 & 0.83 & 27.15 & 0.42 & 317.13 & 0.27 & 19.32 & 1.55 & 340.03 & 0.33 & 18.77 & 2.23\\
    \midrule
    Ours & 21.37 & \textbf{0.926} & \textbf{37.31} & 0.327 & \textbf{170.58} & 0.82 & \textbf{33.68} & \textbf{1.31} & \textbf{121.73} & \textbf{0.83} & \textbf{32.33} & \textbf{1.12} \\
    \bottomrule
  \end{tabular*}
\end{table*}


\section{Experiments}
\label{sec:experiments}


\subsection{Experimental Methodology}
\textbf{Implementation.}
The source diffusion prior of RF-CMG is first pretrained on the Widar3.0 dataset~\cite{zhang2021widar3} and then transferred to self-collected few-shot mmWave and RFID target datasets. The RFID and mmWave datasets contain 8 and 6 action classes, respectively. For mmWave, each class contains 10 samples collected from 10 distinct sensing angles, i.e., one sample per angle. For RFID, samples are collected under the standard acquisition setup. In total, we collect 1,000 samples for each target modality. Only a small subset is used for model adaptation and reference-guided inference, while the remaining real samples are reserved for downstream evaluation; the two subsets are strictly disjoint. We generate 2,000 and 2,500 samples for FID evaluation using a 25-step DDIM~\cite{song2020denoising} sampler.

For the downstream gesture recognition task, we use RF-CMG to generate training data for both target modalities. Specifically, the mmWave dataset contains 6 classes with 2,000 generated training samples and 800 real testing samples, while the RFID dataset contains 8 classes with 2,000 generated training samples and 896 real testing samples. All training samples are generated data, whereas the testing samples are collected from real-world measurements. To avoid the influence of classifier design, we adopt ResNet~\cite{he2016deep} and VGG~\cite{simonyan2014very} as unified evaluators, and train all models for 150 epochs using AdamW with a learning rate of 5e-4, batch size 64, and cosine annealing.

\textbf{Evaluation Metrics.}
We evaluate generation quality using Fréchet Inception Distance (FID), Structural Similarity Index Measure (SSIM), and Peak Signal-to-Noise Ratio (PSNR). FID~\cite{heusel2017gans} measures distribution-level similarity to real data, while SSIM~\cite{wang2004image} and PSNR~\cite{hore2010image} measure structural consistency and pixel-level fidelity against ground-truth targets, respectively. 

For intra-modal generation (WiFi$\rightarrow$WiFi), we follow the standard protocol and report Intra-Learned Perceptual Image Patch Similarity (Intra-LPIPS)~\cite{zhang2018unreasonable, ojha2021few} to measure diversity. For cross-modal generation, however, raw Intra-LPIPS can be misleading, since failed models may produce high-frequency noise that artificially increases diversity scores. 
We therefore introduce \textbf{Relative Intra-LPIPS (r-LPIPS)}:
\begin{equation}
    r\text{-LPIPS} = 
\frac{\text{Intra-LPIPS}_{gen}}
{\text{Intra-LPIPS}_{real}} .
    \label{eq:r-lpips}
\end{equation}
This metric normalizes generated diversity by the intrinsic variability of the target modality. An r-LPIPS value close to $1$ indicates realistic diversity, values much smaller than $1$ suggest mode collapse, and values much larger than $1$ usually indicate artifact-driven diversity inflation.

\textbf{Baselines.} We evaluate our framework against three groups of baselines. To ensure fair comparison, all methods follow the identical source-target split and one-shot target protocol:
(1) \textbf{General Generative Models}: \textbf{DDPM~\cite{ho2020denoising}} and \textbf{Flow Matching~\cite{lipman2022flow}}, serving as fundamental baselines to evaluate standard generative performance.
(2) \textbf{RF-Domain Generative Models}: \textbf{RF-Diffusion~\cite{chi2024rf}}, utilized to examine whether existing wireless-oriented frameworks can inherently handle extreme data scarcity.
Notably, since the methods in these first two groups lack cross-modal transfer capabilities, they are restricted to intra-modality generation directly on the target domain.
(3) \textbf{Few-Shot Adaptation Methods}: State-of-the-art diffusion adaptation techniques, including \textbf{RICK~\cite{zhao2023exploring}}, \textbf{CRDI~\cite{cao2024few}}, \textbf{Uni-DAD~\cite{bahram2025uni}}, and \textbf{DoGFit~\cite{bahram2026dogfit}}. Crucially, because there are no off-the-shelf foundation models dedicated to RF sensing, we implement all these adaptation baselines on top of the exact same pre-trained WiFi DDPM backbone used by our framework. This isolates the effectiveness of the adaptation strategies from the underlying generative prior, thereby guaranteeing a rigorously fair comparison.
\subsection{Overall Generation Quality}
\label{4.2}
The evaluation results for RF-CMG on intra-modal and cross-modal signal generation are presented in Table~\ref{tab:main_results}. 

Despite being designed for cross-modal generation, our method remains competitive in intra-modal WiFi-to-WiFi tasks. It achieves the best SSIM of 0.926 and PSNR of 37.31, while obtaining an FID of 21.37, which is slightly inferior to RF-Diffusion but still outperforms general baselines such as DDPM and Flow Matching. For cross-modal generation, RF-CMG achieves an r-LPIPS of 1.31 for WiFi-to-mmWave and 1.12 for WiFi-to-RFID, significantly outperforming existing baselines. These values are close to the ideal value of 1, indicating that the generated samples exhibit high consistency with real target modalities in terms of both diversity and underlying physical properties. In addition, RF-CMG achieves the lowest FID and highest PSNR on both target modalities, with FID scores of 170.58 for mmWave and 121.73 for RFID. Compared to the best-performing baselines (254.85 for mmWave and 240.03 for RFID), this corresponds to reductions of approximately 33.1\% and 49.3\%, respectively. These results highlight the effectiveness of RF-CMG in leveraging WiFi priors to synthesize high-quality target-modality signals with high fidelity and realistic modality-specific characteristics.

\begin{figure}[t]
    \centering
    \includegraphics[width=\linewidth]{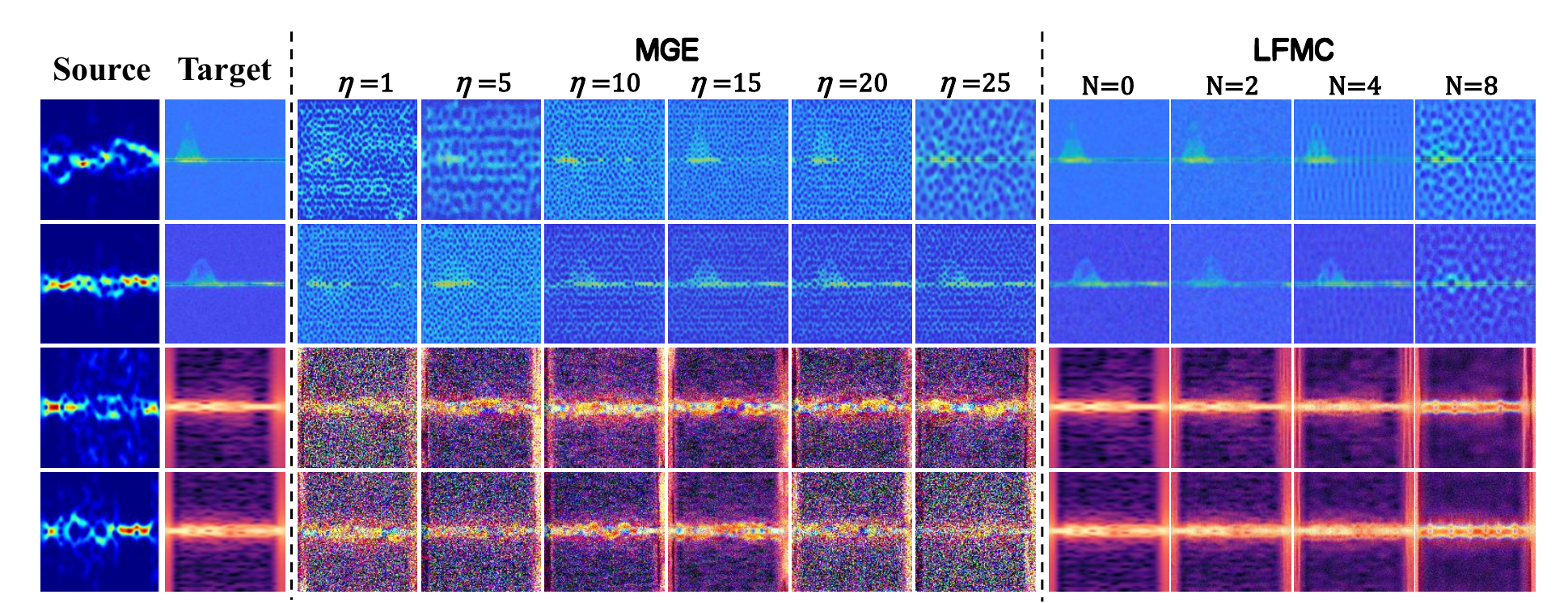}
    \caption{Impact of MGE guidance expressiveness ($\eta$) and LFMC structural strength ($N$) on cross-modal generation. MGE is evaluated without LFMC to isolate high-frequency transfer.}
    \label{fig:Hyperparameter}
\end{figure}
\subsection{Necessity of Balanced Frequency Control}
\label{sec:design-support}
This section showcases an in-depth analysis of the MGE guidance interval $\eta$ and the LFMC scaling factor $N$ within RF-CMG, with the corresponding experimental results illustrated in Fig.~\ref{fig:Hyperparameter}.

\textbf{High-Frequency Guidance ($\eta$).} To isolate the impact of MGE, we remove the LFMC component and modulate only $\eta$. As illustrated in Fig. 6, a small $\eta$ provides insufficient guidance, failing to capture intricate target-modality textures. Conversely, increasing $\eta$ progressively recovers high-frequency patterns, validating MGE’s role in high-frequency adaptation. However, excessively large values ($\eta \geq 20$) lead to overfitting on the sparse target data and exacerbate high-frequency noise. Consequently, we select $\eta=15$ as the optimal trade-off.

\textbf{Low-Frequency Constraint ($N$).}
We then fix $\eta=15$ and vary $N$. The MGE-only results already contain target-modality high-frequency characteristics, but still suffer from noticeable artifacts and unstable global structure. Introducing LFMC with a moderate strength ($N=2$) effectively suppresses these artifacts and preserves the macroscopic layout, whereas overly strong constraints ($N=8$) introduce coarse distortions and harm generation continuity.

The results demonstrate that MGE and LFMC play complementary roles: MGE guides high-frequency detail learning in the target modality, while LFMC regularizes low-frequency structure to mitigate source-modality structural biases and reduce artifacts during generation.


\begin{table}[!t]
  \centering
  \caption{Reference dependency analysis under cross-modal generation.}
  \label{tab:ref_dep}
  \resizebox{\columnwidth}{!}{
    \begin{tabular}{c c c c c }
      \toprule
      & \multicolumn{2}{c}{WiFi $\rightarrow$ mmWave} & \multicolumn{2}{c}{WiFi $\rightarrow$ RFID}  \\
      \cmidrule(lr){2-3} \cmidrule(lr){4-5}
      Condition & SSIM$\uparrow$ & PSNR$\uparrow$ & SSIM$\uparrow$ & PSNR$\uparrow$  \\
      \midrule
      Correct reference & 0.82 & 33.68 & 0.83 & 32.33  \\
      Same class reference  & 0.80 & 31.01 & 0.82 & 31.18  \\
      Cross class reference  & 0.77 & 31.54 & 0.78 & 29.37  \\
      \bottomrule
    \end{tabular}
    }
\end{table}
\begin{figure}[t]
    \centering
    \includegraphics[width=\linewidth]{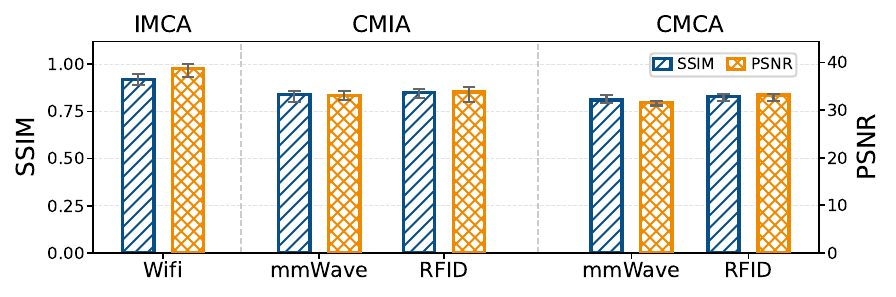}
    \caption{Robust generation evaluation under IMCA, CMIA, and CMCA.}
    \label{fig:generalization_bar}
\end{figure}
\subsection{Reference Sample Sensitivity Analysis}
To examine the sensitivity of LFMC to the reference sample choice, we fix the same noisy input $x_t$ and reconstruction target $x_0$, and vary only the reference used at inference: (i) \emph{correct} reference sample $x_0$, (ii) a reference sample from the same gesture class but a different instance, and (iii) a reference sample from a different class (Table~\ref{tab:ref_dep}). This process does not involve retraining. As shown in Table~\ref{tab:ref_dep}, the correct reference yields the best reconstruction performance, while both intra-class (different-sample) and cross-class references incur only moderate degradation. This indicates that LFMC benefits from compatible reference guidance without being restricted to any specific reference instance.

\subsection{Extreme Generalization Analysis}
\label{subsec:generalization}
We evaluate generalization under three progressively harder settings: Intra-Modal Cross-Action (IMCA), which synthesizes unseen actions within the source WiFi modality; Cross-Modal Intra-Action (CMIA), which translates a seen action to the target modality; and Cross-Modal Cross-Action (CMCA), which synthesizes unseen actions in the target modality with limited one-shot guidance.

As shown in Fig.~\ref{fig:generalization_bar}, performance drops from IMCA to CMIA due to the modality gap, but degrades only slightly further under the more challenging CMCA setting. Specifically, compared with CMIA, CMCA reduces SSIM by only 0.029 on mmWave and 0.022 on RFID, while PSNR drops by 1.51 dB and 0.61 dB, respectively. Even in this most stringent setting, RF-CMG still achieves SSIM values of 0.811 and 0.828 on the two target modalities. These results indicate that the frequency-decoupled design generalizes beyond seen target instances, where MGE captures transferable high-frequency target cues and LFMC preserves stable low-frequency structure.

\begin{figure}[t]
    \centering
    \includegraphics[width=\linewidth]{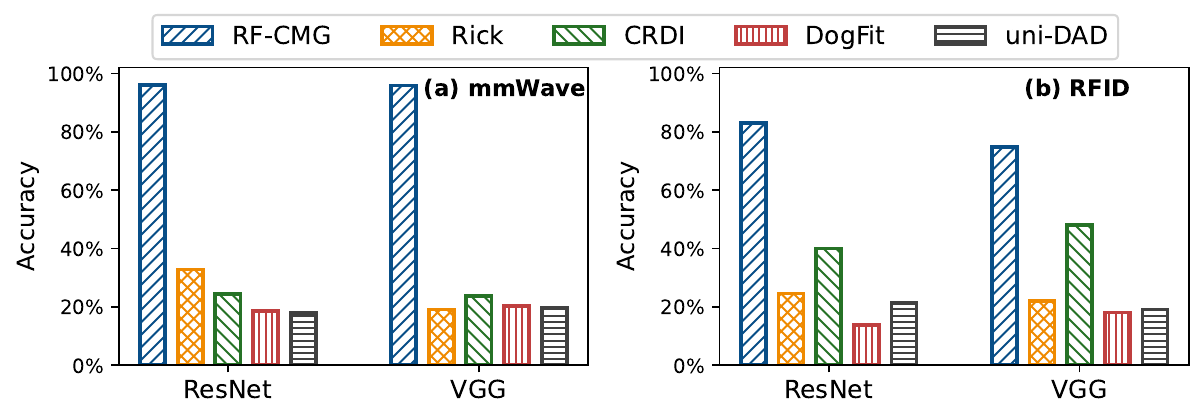}
    \caption{Performance comparison of downstream gesture recognition.}
    \label{fig:downstream_gesture}
\end{figure}

\begin{figure}[t]
    \centering
    \includegraphics[width=\linewidth]{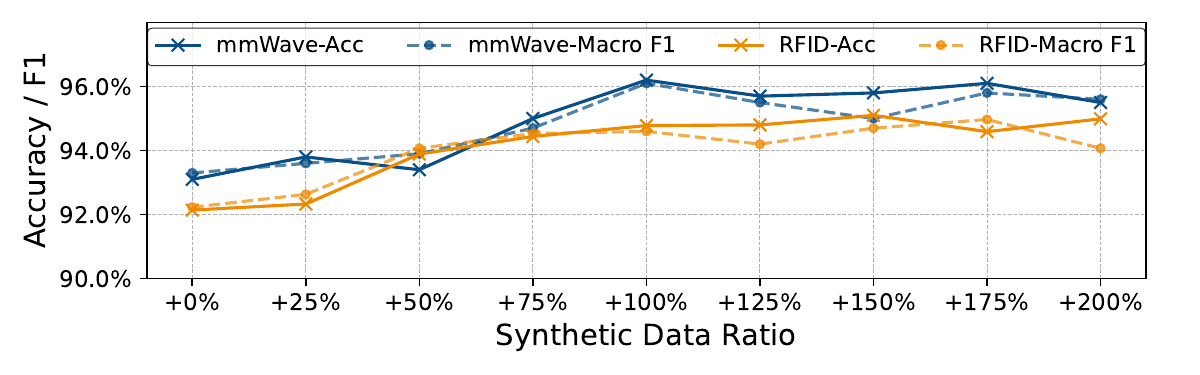}
    \caption{Impact of the synthesized-data ratio on gesture recognition performance.}
    \label{fig:syn_ratio}
\end{figure}
\subsection{Downstream Applications}

\textbf{Gesture Recognition.} As shown in Fig.~\ref{fig:downstream_gesture}, when utilizing synthesized samples to train downstream classifiers, our frequency-decoupled framework demonstrates a substantial performance advantage over existing baselines. For mmWave, RF-CMG achieves 96.02\% and 95.85\% accuracy with ResNet and VGG, surpassing the baselines by 63.24\% and 72.19\%, respectively. For RFID, RF-CMG attains 83.10\% accuracy with ResNet, compared to 39.96\% for the baseline method. These results indicate that RF-CMG can generate physically meaningful target-modality features and substantially narrows the sim-to-real gap under extreme data scarcity.



\textbf{Impact of Synthesized Data Ratio.}
We fix the number of real training samples to 1,000 for mmWave and 700 for RFID, and progressively augment the training set with synthesized data at ratios ($r \in \{0, 0.25, \dots, 2.0\}$) to evaluate the impact of synthesized data. The corresponding real test sets contain 300 and 180 samples, respectively. As shown in Fig.~\ref{fig:syn_ratio}, both target modalities exhibit a similar trend, where downstream performance improves as more synthesized data is introduced and then gradually saturates. This saturation occurs earlier for mmWave, around +100\%, whereas RFID continues to benefit up to a higher range of approximately +150\% to +175\%. These results indicate that synthesized cross-modal data consistently improves downstream recognition, while the optimal mixing ratio remains modality-dependent.
\begin{table}[!t]
  \centering
  \caption{Comparisons of model performance from one-shot to few-shot, evaluated by the FID ($\downarrow$).}
  \resizebox{\columnwidth}{!}{
    \begin{tabular}{l c c c c c c}
      \toprule
      & \multicolumn{2}{c}{1-shot} & \multicolumn{2}{c}{5-shot} & \multicolumn{2}{c}{10-shot} \\
      \cmidrule(lr){2-3} \cmidrule(lr){4-5} \cmidrule(lr){6-7}
      Methods & mmWave & RFID & mmWave & RFID & mmWave & RFID  \\
      \midrule
      RICK & 275.34 & 271.22 & 225.47 & 195.44 & 153.17 & 136.74 \\
      CRDI  & 254.85 & 240.03 & 219.32 & 205.12 & 201.78 & 192.31 \\
      DoGFit  & 296.29 & 253.93 & 244.77 & 201.23 & 203.71 & 195.64 \\
      Uni-DAD  & 314.13 & 340.03 & 284.64 & 293.51 & 184.83 & 171.72 \\
      Ours  & \textbf{170.58} & \textbf{121.73} & \textbf{155.87} & \textbf{107.38} & \textbf{142.52} & \textbf{91.62} \\
      \bottomrule
    \end{tabular}
    }
    \label{tab:few}
\end{table}
\begin{figure*}[!t] 
  \centering
  \includegraphics[width=\textwidth]{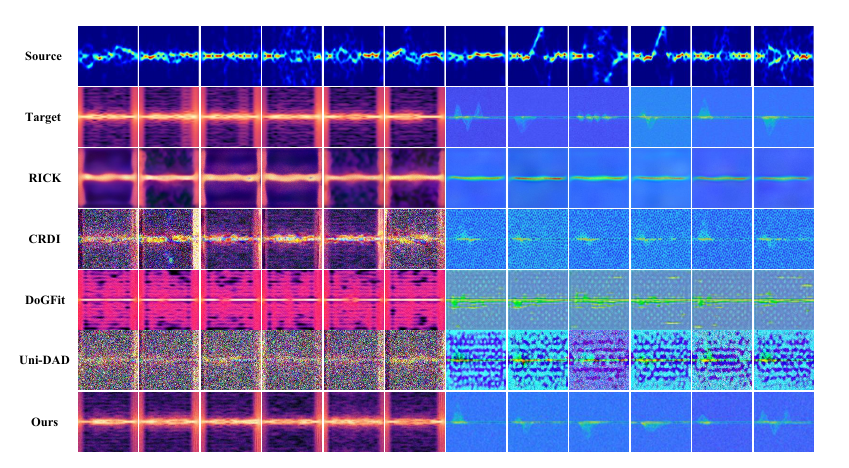}
  \caption{Visualization of cross-modal generation results across baseline methods and RF-CMG.}
  \label{fig:qualitative}
\end{figure*}
\subsection{From One-Shot to Few-Shot}
\label{subsec:Few-Shot}
Since mmWave signals are highly sensitive to sensing angle, our 1-shot setting for mmWave denotes one reference sample per angle (10 angles in total) for each action class, rather than a single pooled exemplar. For RFID, which does not use angle-specific partitioning, $K$-shot follows the standard definition of $K$ target samples per class. To study data scalability, we increase the number of target references to $K \in \{5,10\}$ and report the resulting FID in Table~\ref{tab:few}. As expected, more target samples improve all methods; however, RF-CMG consistently achieves the lowest FID on both mmWave and RFID across all $K$, showing that the proposed frequency-decoupled design remains effective beyond the strict few-shot regime.



\subsection{Qualitative Evaluation}
\label{subsec:qualitative}
To provide a more intuitive comparison of cross-modal generation quality, we visualize examples generated by RF-CMG and baseline methods in Fig.~\ref{fig:qualitative}. RF-CMG produces cleaner structural patterns and more realistic target-modality details. The synthesized samples exhibit stronger physical plausibility, making them more suitable for supporting downstream recognition tasks. In contrast, RICK attempts to transfer knowledge via progressive weight adaptation. However, the substantial modality gap from WiFi to mmWave and RFID leads to over-smoothed, blob-like artifacts in the generated results. CRDI improves diversity via conditional relaxed diffusion inversion but fails to bridge structural discrepancies, resulting in incomplete target patterns contaminated by source-domain noise. Methods such as DoGFit and Uni-DAD, which rely on domain-guided fine-tuning and dual-domain distillation, tend to collapse under the large cross-modal gap, producing high-frequency artifacts instead of meaningful structures.



\section{Conclusion}
\label{sec:conclusion}
This paper presents RF-CMG, the first cross-modal generative diffusion model designed for RF signals. RF-CMG excels in generating high-fidelity millimeter-wave and RFID data from low-cost WiFi signals by decoupling the complex cross-modal generation process into high-frequency guidance and low-frequency constraint stages. The MGE module explicitly guides the synthesis of high-frequency, modality-specific textures of the target modality, while the LFMC module progressively imposes low-frequency constraints during inference. This design effectively suppresses source-modality structural biases while preserving the fundamental generative dynamics of the pre-trained backbone. RF-CMG exhibits remarkable performance in cross-modal generation tasks, and the generated data demonstrates significant potential in gesture recognition, shedding new light on the application of AIGC in wireless research.

\bibliographystyle{ACM-Reference-Format}
\bibliography{sample-base}

\appendix









\end{document}